\documentclass{article}
\usepackage{ijcai24}

\usepackage{times}
\usepackage{soul}
\usepackage{url}
\usepackage[hidelinks]{hyperref}
\usepackage[utf8]{inputenc}
\usepackage[small]{caption}
\usepackage{graphicx}
\usepackage{amsmath}
\usepackage{amsthm}
\usepackage{amssymb} 
\usepackage{booktabs}
\usepackage{algorithm}
\usepackage{algorithmic}
\usepackage[switch]{lineno}
\usepackage{xcolor}

\title{Automate Knowledge Concept Tagging on Math Questions with LLMs}

\author{
Hang Li$^{1,2}$
\and
Tianlong Xu$^1$\and
Jiliang Tang$^2$\And
Qingsong Wen$^1$\footnote{Corresponding author}\\
\affiliations
$^1$Squirrel AI, Bellevue, WA, USA\\
$^2$Michigan State University, USA\\
\emails
\{lihang4, tangjili\}@msu.com,
\{tianlongxu, qingsongwen\}@squirrelai.com
}

\begin{document}

\maketitle

\begin{abstract}
Knowledge concept tagging for questions plays a crucial role in contemporary intelligent educational applications, including learning progress diagnosis, practice question recommendations, and course content organization. Traditionally, these annotations have been conducted manually with help from pedagogical experts, as the task requires not only a strong semantic understanding of both question stems and knowledge definitions but also deep insights into connecting question-solving logic with corresponding knowledge concepts. In this paper, we explore automating the tagging task using Large Language Models (LLMs), in response to the inability of prior manual methods to meet the rapidly growing demand for concept tagging in questions posed by advanced educational applications. Moreover, the zero/few-shot learning capability of LLMs makes them well-suited for application in educational scenarios, which often face challenges in collecting large-scale, expertise-annotated datasets. By conducting extensive experiments with a variety of representative LLMs, we demonstrate that LLMs are a promising tool for concept tagging in math questions. Furthermore, through case studies examining the results from different LLMs, we draw some empirical conclusions about the key factors for success in applying LLMs to the automatic concept tagging task.











\end{abstract}

\section{Introduction}


Knowledge concept tagging, which targets at generate precised knowledge index to educational materials, has been recognized as an important factor of current intelligent education systems in providing high-quality educational contents to both educators and learners during the practice \cite{chen2014tag}. For example, with well-annotated education materials, teachers will receive great conveniences in organizing coursing contents through searching concept key words index \cite{sun2018automatic}. Among the tagging objects, concept tagging over math questions have been greatly emphasised because of the recent successes of applying intelligent tutoring system (ITS) in mathematical education \cite{burns2013intelligent}. Traditionally, the questions' concept tags are annotated by the pedagogical expertise. However, the rapid growth of the internet has caused that conventional manual methods insufficient to meet the demand for handling large volumes of online question data or updating exist concept tags in a timely fashion. In addition, the size limitation of annotated samples also impedes the wide application of deep learning methods in ITS. 

In order to solve the above issues, many pioneering researches have tried to automate the tagging process with different natural language processing (NLP) algorithms \cite{sun2018automatic,zhang2021question}. For example, early works use text embedding techniques to convert the text contents of questions into dense vectors, train machine learning models, and then classify them into predefined tag categories \cite{du2021application}. However, these practices overlook the vital relationship in solutions and knowledge concepts, which leads to unsatisfactory tagging results. Recent studies solve the problem and improve the tagging performance by leveraging pre-trained language models (PLMs) to fuse external information, such as solution text and conceptual ontology, with original question contents \cite{huang2023pqsct}. Unfortunately, these new trials introduce additional input data requirements to the knowledge tagging model, which restricts the wide applications of the algorithm to question with limited external resources. In order to solve the above challenges and keep the algorithm scalable to wider question concept tagging cases, we leverage LLMs as backbone models for the automatic knowledge tagging problem.

Leveraging the advanced mathematical and logical inference capabilities inherent in Large Language Models (LLMs), our method eliminates the need for external information, such as solution text, as a mandatory input to link solution-related knowledge concepts with the given question content. Furthermore, LLMs can dynamically generate solutions related to concepts, offering more precise tagging results compared to the static solution texts explored in previous research. Additionally, owing to the strong zero-shot or few-shot learning abilities of LLMs, our approach can be swiftly applied with minimal annotation samples, setting it apart from all previous training-based algorithms. This feature allows our method to be rapidly adapted for annotating works encompassing nearly all knowledge concepts and questions. To validate the effectiveness of our proposed framework, we have collected over 1,000 expertly annotated concept tagging samples as a test set. Through extensive experiments across various LLMs, we have shown that employing LLMs with appropriate prompt tuning techniques holds promise as tools for concept tagging. Finally, by comparing the results from different LLMs, we have drawn some empirical insights into the crucial factors that ensure success in using LLMs for automated knowledge concept tagging tasks.

\section{Related Work}

\subsection{Knowledge Concept Tagging}

The major challenge of the tagging tasks is how to construct a meaningful link in between the knowledge concepts and the problems, either through the description of the problem themselves or through solutions. The formulation of the task can primarily be categorized into two directions: retrieval and matrix decomposition. The former relies heavily on training a semantic representation. \cite{8295250} employs simple backbone models such as long short term memory (LSTM) and some attention mechanisms to learn short-range dependency embeddings, where the questions are fed into LSTM layers and are ultimately connected to cross entropy functions that indicate whether or not a tagging concept belongs to a given problem. \cite{liu2019ekt} devised an exercise-enhanced recurrent neural network with Markov property and Attention mechanism to extract rich knowledge concepts information in the exercise's content. Similarly but with enriched data source such as text,  multi-modal data \cite{yin2019quesnet} as well as latex formula combined data \cite{huang2021context}, semantic representations learned with LSTM have been improved to capture more implicit contexts. \cite{huang2020neural} fills knowledge graph information into the embedding layers and achieves better mathematical semantic understanding. To take advantage of the robust transformers framework, \cite{zemlyanskiy2021docent} pretrained a BERT model to learn jointly predicting words and entities as movie tags given the reviews of movies. \cite{10123979} proposes an improved pretrained bidirectional encoder representation from transformers (BERT) for concept tagging with both questions and solutions. This work formulates a next-sentence prediction with question–solution that fits into the BERT encoder, ultimately obtaining the final exercise representation through feature augmentation. The work also conducts a pseudo-siamese training manner to learn the hidden mapping relationships between questions and concepts in mathematics specifically. The latter mostly applies to relatively static concept-problem set (i.e., neither the concept nor the problem change drastically). The can be formulated as a sparse matrix $Q \in \mathbb{R}^{N \times K}$ where N and K represent the number of problems and number of concepts respectively. Major techniques include matrix factorization \cite{desmarais2013matrix} and Bayesian estimation \cite{chen2018bayesian}, etc. Similarly to retrieval, such Q matrix can aggregate more abundant features \cite{huang2019hierarchical} to augment the power of Q matrix estimation.

\subsection{Annotation Tasks with LLMs}

With the evolving of LLMs and their strong performance on dealing with unstructured data, some attempts have been leveraging LLMs to directly produce annotations in the question-answering manner. \cite{dong2024language} leverages GPT series and fine-tuned FLAN-T5 \& Llama 2 to insert extracted concepts into a predefined concept ontology graph. \cite{yan2023biomedical} injected the biomedical domain knowledge graph into LLM by pre-training the model such that difficulties of linking the biomedical entities with LLMs are tackled. \cite{bacciu2023rraml} created a reinforcement learning retrieval augmented machine learning framework to enhance LLM's performance for wider range of relevant entity retrieval. \cite{wang2023exploring} performs in-context learning methods and used LLM-ranking to link bio-medical entities (i.e., what's mentioned as two-stage retrieve and rank framework).  \cite{dunn2022structured} finetuned GPT3 with around 500 pairs of prompt-expected json format entities, and achieved meaningful scientific entities annotation(subjects include chemistry, phase, and morphology) in a hierarchical structure. \cite{xiao2023instructed} proposed an instructed generative entity linker for precise entity predictions over a large knowledge base, which is composed of a lightweight potential mention retriever and sequence-to-sequence training EL objective with instruction-tuning. \cite{masson2023optimal} demonstrates that few-shot learning is superior over fine-tuning on LLMs while performing the thematic concept annotation task in the tourism domain when, in their specific case, the quantity of training data is limited. \cite{lin2024gumsley} introduced a novel salient entity annotation (entities that are central to a document’s overall meaning) approach that enhances LLMs' understanding of the underlying semantics of English sentences. \cite{ding2024chatel} formulated a three-step chat entity linking framework to map hidden relationships between concepts and sentence texts: (1) given a sentence, process entity candidate generation step to obtain relevant entities. Then (2) an augmentation step is performed to obtain an auxiliary content of the annotated mention. Finally (3) a multi-choice selection prompt is conducted to decide the corresponding entity of annotated mention.

Overall, leveraging LLMs to perform concept annotations primarily relies on nuanced few-shot prompting, fine-tuning the model, as well as providing extra meaningful data augmentations. The major challenges and opportunities are how to build a semantic link between the text contents and concepts, particularly when only little or none of the domain knowledge is known to LLMs.

\section{Methodology}
\subsection{Overview}


The problem of knowledge concept tagging can be articulated as: Given a specific knowledge concept $k_i$ from a set $K = \{k_1,...,k_m\}$ and a question $q_j$ from a set $Q=\{q_1,...,q_n\}$, the objective of a concept tagging model is to produce a binary judgment $y\in\{0,1\}$. This judgment indicates whether or not $q_j$ aligns with $k_i$. Previous studies have approached this by encoding both the question and the knowledge text into dense vectors using various embedding models. Following this, a classifier, typically trained with binary cross-entropy loss, is employed to generate the binary judgment outcomes.

\begin{figure}
    \centering
    \includegraphics[width=0.48\textwidth]{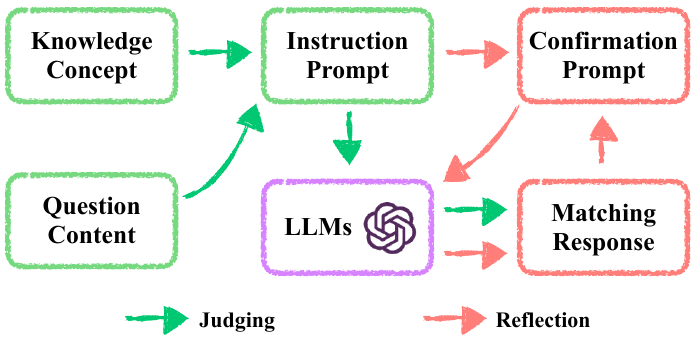}
    \caption{Overview of our purposed framework.}
    \label{fig:framework}
\end{figure}

In this paper, we introduce an approach that capitalizes on the zero-shot and few-shot learning capabilities of Large Language Models (LLMs), thus eliminating the need for task-specific training data to fine-tune the model's parameters. Our method involves crafting suitable prompt texts that instruct the LLMs on the objective of the concept tagging task. We determine the judgment by analyzing the positive or negative sentiments expressed in the responses generated by LLMs. To enhance the performance of LLMs on the concept tagging task further, we have developed various prompt optimization strategies. An overview of our proposed framework is depicted in Fig~\ref{fig:framework}.

\subsection{Zero-shot Prompt Designs}

\label{sec:zeroshot}

The most significant difference between LLMs and other prior machine learning models is its zero-shot learning capability \cite{wei2021finetuned}. Contributing to the huge size model parameter and the extensive pre-training on diverse and vast dataset, LLMs have demonstrated on their strengths in comprehending instructions in natural language and applying learned knowledge to new problems without requiring additional training data specific to these tasks. In our case, we first describe the goal of the tagging task as: \emph{You are a knowledge concept annotator. Your job is to judge whether the given Knowledge is matching the Question}. For the convenience of following process, we also add a response format instruction in the prompt: \emph{Your answer should start with 'Yes' or 'No'.} After that, as the prior studies like Chain-of-Thought (COT) \cite{wei2022chain} have discovered that with instructing LLMs to generate step-by-step problem solving solutions will be helpful for the LLMs to draw the correct conclusions while facing some complicated problems. Therefore, we ask LLMs to not only provide its positive or negative sentiments but also present the reason, \emph{You should also provide your reason for your judgment.} At last, since all the question in our test set will only be connected with their own correct knowledge concept, we introduce the prior knowledge to LLMs through the instruction: \emph{If Question covers other Knowledge, your answer should be 'No'.} Overall, the zero-shot task instruction prompt is presented as follows:

\begin{quote}

\emph{\textbf{Instruction}: You are a knowledge concept annotator. 
Your job is to judge whether the given Knowledge is matching the Question. 
Your answer should start with 'Yes' or 'No'. 
You should also provide your reason for your judgement. 
If Question covers other Knowledge, your answer should be 'No'.}

\emph{\textbf{Knowledge}:} \underline{The composition of numbers within 20.}

\emph{\textbf{Question}:} \underline{There are (\ ) tens and (\ ) ones in 14.}

\emph{\textbf{Judgement}:} (Generated by LLMs)

\end{quote}

The text marked with underscore is the knowledge text $k_i$ and question text $q_j$. During the practice, we notice that using the original knowledge concept names directly usually yields sub-optimal results, since some descriptions are too ambiguous. To conquer this challenge, we asks expertise to give some specific definitions to those unclear concepts and mark each the covering boundaries of each knowledge concept. With these modifications to the definition of the knowledge concepts, LLMs can achieve satisfying tagging results, and detailed comparison result can be found in Appendix. The detailed knowledge definitions and covering boundary of each knowledge concept are shown in table~\ref{tab:data_detail}.

\subsection{Few-shot Demonstration Selection}

Although the zero-shot prompt provides a good solution without using any annotated samples, the brief description text sometimes may not be specific enough to cover some complicated cases. For example, there is a knowledge concept named \textbf{consecutive carry in multiplication}, which occurs when the product of two digits, along with any carry from the previous calculation, results in a number greater than 9, thus requiring another carry to be added to the next column in the calculation. It will be easier for LLMs to understand and follow the instruction by presenting some demonstration samples, e.g., $38 \times 9$, and their corresponding explaining text: 
\begin{quote}
    \textit{There is a consecutive carry starts from the multiplication of the ones place ($8 \times 9 = 72$, carry $7$), and then a carry from the tens place operation ($3 \times 9 + 7 = 34$, carry $3$).}
\end{quote}

Apart from that, through presenting the example answers as the inputs and instruct LLMs to generate similar form of responses, LLMs will learn and follow the judging steps of demonstrations, which helps it output more relevant responses. We show the comparison details between zero-shot and few-shot responses in Sec.~\ref{sec:fewshot}. During the implementation of few-shot learning prompts, many recent studies found that the final outcomes of LLMs are significantly influenced by the selected demonstration samples \cite{wei2023larger} and many demonstration selecting algorithms have been proposed to fully exploit the potentials of few-shot learning capability of LLMs \cite{margatina2023active,su2022selective}. However, since we focus on implementing zero-shot and few-shot learning for the knowledge concept tagging task, we do not include the advanced demonstration recommending algorithms, which has the requirements for annotated samples, in this paper. We leave the exploration for this direction in our future work. Following the empirical conclusions drawn from the prior works \cite{su2022selective}, we explore three heuristic-based demonstration selecting strategies to enhance the few-shot learning results. 

\begin{itemize}
    \item \textbf{Knowledge Relevance:} 
    In this strategy, we consider the relevance between the knowledge concept definition text $k_i$ of different samples $x_i = (k_i, q_i) \in \mathcal{X}$, where $\mathcal{X}$ is the test sample set. By selecting the samples sharing the same of similar $k_i$ as demonstrations, we hope the LLMs can follow the correct judging steps and generate the accurate results.
    \item \textbf{Question Diversity:}
    The idea behind this strategy is inspired by prior work \cite{mavromatis2023examples}, which demonstrates that through including diverse few-shot samples, LLMs present stronger general capabilities while facing some margin samples during the test. In our work, we use the sentence-level embedding generated by LLMs and perform K-means clustering to find the diverse cluster groups within the question representation manifold. Finally, we choose one question for each cluster, which is closest to the cluster center, as demonstrations.
    \item \textbf{Label Distribution:}
    The label distribution of demonstration samples could also influence the performance of few-shot learning responses of LLMs, as different categories of labels provide support information from their own perspectives \cite{yao2023more}. For example, in this paper, the match demonstrations ($y_i = 1)$ present LLMs the correct question example $q_i$ which matches all the requirement of the given knowledge $k_i$. And the mismatch samples ($y_i = 0)$ could display the common errors of the task and help LLMs to avoid making similar errors during the generation.  
\end{itemize}






\subsection{Self-reflection Revision}

A class of methods that leverages LLMs’ ability to self-reflect based on feedback from the environment has shown their superiority compared to algorithms that do not have an awareness of doing the task a second time \cite{zhao2023expel}. In our framework, we ask LLMs to make a deterministic statement given the original instructional prompt and the last round responses generated by LLMs. During the implementation, we found it essential to tune the deterministic instruction prompt into a neutral tone. For example, we try to avoid using some biased expression such as "please double check", "are you sure", otherwise the LLMs will be misguided to always giving reverse judgement to the original response. Here is an example of our self-reflection prompt for the later experiment.

\begin{quote}
    \emph{(Start with the instruction, knowledge and question prompt presented in Sec~\ref{sec:zeroshot})}

    \emph{\textbf{Judgement}:} \ul{Yes, the knowledge matches the question. The question is asking for the number of tens and ones in a given two-digit number (14) which falls within the range specified in the knowledge (up to 20, inclusive).}

    \emph{\textbf{Instruction}: Check the knowledge and question and confirm whether the prior answer is correct or wrong.} 

    \emph{\textbf{Confirmation}:} (Generated by LLMs)
\end{quote}

The text marked with underscore is the last response text generated by LLMs. In addition, as concept tagging results are used as the searching index for many downstream education applications, the requirement to the the precision of generated tagging results usually receive more attentions compared to the recall. Based on this fact, we apply the self-reflection revision only to the test samples which receive the positive response. By incorporating such strategy, our framework not only achieve a much higher precision, but also receive a huge boost over the execution efficiency.



\section{Experiment}
\subsection{Dataset Overview}

In order to testify the effectiveness of our proposed framework, we collect a knowledge concept tagging dataset, MathKnowCT, from the online K-12 math education materials. The dataset contains 12 knowledge concepts, spreading from the study requirements of grade 1 to grade 3 students in elementary schools. For each knowledge concept, we collect more than 80 candidate questions and ask a pedagogical expert to annotate whether the question and knowledge pair is match. The ratio between matching and mismatching categories is 1:8. More details about the dataset statistics and knowledge concept definitions can be found in Table~\ref{tab:data_detail}.

\begin{table*}[]
\caption{Details of knowledge concept definitions and dataset statistics.}
\label{tab:data_detail}
\resizebox{.95\textwidth}{!}{
\begin{tabular}{@{}lllll@{}}
    \toprule
    Knowledge   Concept & Definition \& Boundary & Number of Questions & Number of Match & Number of Mismatch \\ \midrule
    \begin{tabular}[c]{@{}l@{}}The composition of \\ numbers within 20\end{tabular} & \begin{tabular}[c]{@{}l@{}}Directly give a two-digit number within 20 \\ (including 20) and read out how many tens or \\ ones it contains, or give several tens and several ones \\ and fill in the corresponding two-digit number. \\ There are no pictures to represent numbers, \\ and numbers above 20 and   above are not involved.\end{tabular} & 117 & 15 & 102 \\ \midrule
    \begin{tabular}[c]{@{}l@{}}Comparison of \\ numbers within 5\end{tabular} & \begin{tabular}[c]{@{}l@{}}Comparison between integer numbers within 5, \\ which can be between two numbers or arranged \\ in order of   numerical value; the question stem \\ needs to provide specific numbers or   present \\ numbers within a situational description; a \\ comparison of numerical   values is required. \\ Does not include calculations.\end{tabular} & 116 & 13 & 103 \\ \midrule
    Reciprocal & \begin{tabular}[c]{@{}l@{}}Directly write the reciprocal of a fraction, \\ an integer or a decimal. Otherwise, find the \\ number whose product with a given number is 1. \\ no letters involved.\end{tabular} & 109 & 26 & 83 \\ \midrule
    Use of reciprocals & \begin{tabular}[c]{@{}l@{}}Given two or three algebraic expressions (where \\ the expressions are in the form of a number multiplied \\ by a single letter) are equal, assuming the value is 1, \\ use the method of finding a number's reciprocal to \\ determine and compare the values of the letters;   \\ or if it is known that two algebraic expressions are \\ equal and their value is 1, find the value of the letters \\ and then substitute this value into another expression \\ to calculate its value. The problem must include letters. \\ The letter index can only be 1, and does not involve \\ squares and other higher orders.\end{tabular} & 148 & 9 & 139 \\ \midrule
    \begin{tabular}[c]{@{}l@{}}Use the properties of \\ decimals to simplify \\ decimals\end{tabular} & \begin{tabular}[c]{@{}l@{}}Utilize the properties of decimals (adding 0 to the end \\ of a decimal or removing 0 does not change the value \\ of the decimal) to simplify decimals. The requirement \\ is to simplify to the simplest form, meaning there are \\ no longer any zeros at the end of the decimal part. \\ Without calculation.\end{tabular} & 71 & 13 & 58 \\ \midrule
    \begin{tabular}[c]{@{}l@{}}Use the properties of \\ decimals to rewrite \\ decimals\end{tabular} & \begin{tabular}[c]{@{}l@{}}Utilize the properties of decimals (adding 0 to the end \\ or removing 0 from the end of a decimal does not change \\ its value) to rewrite decimals as required. The requirement \\ is to rewrite them as decimals with a specific number of \\ decimal places (one, two,   three, etc.) or as decimals in \\ units of one-tenth, one-hundredth, or one-thousandth. \\ Without calculation\end{tabular} & 113 & 19 & 94 \\ \midrule
    \begin{tabular}[c]{@{}l@{}}The rules of decimal \\ point movement\\ -calculation\end{tabular} & \begin{tabular}[c]{@{}l@{}}Obtain the calculation result based on the movement of \\ the decimal point when a decimal is multiplied by 10, 100, \\ 1000, or divided by 10, 100, or 1000; or, given the result of \\ the decimal point movement, determine whether the decimal \\ was multiplied or divided by a certain number (10, 100, or \\ 1000); all calculations are done in one step. No two-step \\ calculations or multiple moves\end{tabular} & 101 & 17 & 84 \\ \midrule
    \begin{tabular}[c]{@{}l@{}}Understanding and \\ Classification of \\ Numbers\end{tabular} & \begin{tabular}[c]{@{}l@{}}Understanding of numbers, including the definitions of integers, \\ fractions, decimals, positive numbers, negative numbers, \\ and natural numbers; needing to select numbers of a specified \\ type from a set of numbers; determining whether a number falls \\ within a defined range; judging the truth or falsehood of \\ propositions related to the classification of numbers. \\ Does not include number operations and irrational numbers;\end{tabular} & 95 & 30 & 65 \\ \midrule
    \begin{tabular}[c]{@{}l@{}}Adding and subtracting \\ whole tens - word problems\end{tabular} & \begin{tabular}[c]{@{}l@{}}One-step application problems involving the addition of \\ whole tens to whole tens or the subtraction of whole tens \\ from whole tens; the calculation result is less than 100. \\ There are no whole hundreds, only two-digit whole tens.\end{tabular} & 142 & 4 & 138 \\ \midrule
    \begin{tabular}[c]{@{}l@{}}Adding and subtracting \\ integer tens - \\ Comparison of formulas\end{tabular} & \begin{tabular}[c]{@{}l@{}}Comparing the result of adding two whole tens with another \\ number; or comparing the size of expressions involving the \\ addition or subtraction of two whole tens; the calculation \\ requires a one-step addition or subtraction of two whole tens; \\ there needs to   be a step for comparing sizes. There are no \\ whole hundreds, only two-digit whole tens.\end{tabular} & 97 & 11 & 86 \\ \midrule
    \begin{tabular}[c]{@{}l@{}}Two-step operation of \\ adding and subtracting \\ integer tens\end{tabular} & \begin{tabular}[c]{@{}l@{}}Calculate the result of adding and subtracting three whole tens; \\ or first present the addition and subtraction of three single-digit \\ numbers, then give the result of adding and subtracting the \\ corresponding whole tens of these three single-digit numbers;   \\ the result needs to be within 100. There are no whole hundreds, \\ only two-digit whole tens.\end{tabular} & 103 & 10 & 93 \\ \midrule
    \begin{tabular}[c]{@{}l@{}}Area unit--unit \\ conversion\end{tabular} & \begin{tabular}[c]{@{}l@{}}Unit conversion between square meters, square decimeters, \\ square centimeters, and square millimeters; conversions are \\ made directly according to the conversion rate between area \\ units, including conversions from larger units to smaller units \\ and from smaller   units to larger units. Does not include word \\ problems; does not include   square kilometers\end{tabular} & 147 & 19 & 128 \\ \midrule
    Overall &  & 1359 & 186 & 1173 \\ \bottomrule
    \end{tabular}}
\end{table*}

\subsection{Experiment Settings}

To fully exploit the potentials of leveraging LLMs for knowledge concept tagging task, we experiment with 5 representative LLMs, including GPT \cite{brown2020language}, LLAMA2\cite{touvron2023llama}, Mixtral \cite{jiang2024mixtral}, Qwen1.5 \cite{bai2023qwen}, and InternLM2 \cite{ying2024internlm}. To be noticed, in this paper, we only experiment with each LLM's instruct-tuned version, since we observe that the instruct-tuned LLMs can better follow the given instruction and generate the correct format responses. When conducting zero-shot or few-shot learning with the raw versions, LLMs will commonly generate question solutions instead of the matching judgments. To convert the generated text response into the binary judgements, we use the regular expression to find the positive and negative pattern of the text. The evaluation metrics used in our experiments are accuracy, precision, and recall. We use 8*40G Nvidia A100 GPUs for all of our experiments. And the LLMs are implemented with huggingface packages \footnote{\url{https://huggingface.co/}}. 

\subsection{Zero-Shot Learning Results}
\label{sec:zero-shot}

We first experiment with the zero-shot performance of LLMs over the different prompt designs and the results are shown as table~\ref{tab:zero-shot}. From the table, we observe that by refining the task instructing prompt text with additional output format and reasons requests ($\mathrm{Instruct\ Prompt\_v2}$), the performance of LLMs is improved, which demonstrates the effectiveness of our prompt optimizations. On the other hand, after enriching the original knowledge concept terms with specific definition words and knowledge boundary descriptions, all LLMs receive another round of significant boost in their performance, which also indicates the importance of the knowledge concept interpretations. At last, by comparing different LLMs, we find the performance gaps between different LLMs are quite large. For example, the most advanced LLMs such as GPT-4 can achieve nearly 90\% accuracy even under the zero-shot scenario, while some other open-source LLMs, such as LLAMA2-70B-Chat, only achieves around 60\% result. 

\begin{table}[]
\caption{Evaluation result of LLMs under zero-shot settings. The best result is marked with \textbf{bold}, the second best is marked with \underline{underline}.}
\label{tab:zero-shot}
\resizebox{.48\textwidth}{!}{
\begin{tabular}{@{}ccccc@{}}
\toprule
\multicolumn{1}{c|}{\textbf{Model}} & \textbf{Accuracy} & \textbf{Precision} &\ \ \  \textbf{Recall}\ \ \ &\ \ \ \ \ \textbf{F1} \ \ \ \ \\ \midrule
\multicolumn{5}{c}{\textbf{Instruction   Prompt\_v1}} \\ \midrule
\multicolumn{1}{c|}{GPT-3.5-Turbo} & \underline{.8433} & .2319 & .4215 & .2992 \\
\multicolumn{1}{c|}{GPT-4} & \textbf{.8792} & \textbf{.4079} & .9430 & \textbf{.5695} \\
\multicolumn{1}{c|}{LLAMA2-70B-Chat} & .5812 & .1603 & \underline{.9437} & .2740 \\
\multicolumn{1}{c|}{Mixtral-8*7B-Instruct} & .7661 & \underline{.2468} & .8811 & \underline{.3856} \\
\multicolumn{1}{c|}{Qwen1.5-72B-Chat} & .7477 & .2413 & .9365 & .3837 \\
\multicolumn{1}{c|}{InternLM2-20B-Chat} & .6063 & .1691 & \textbf{.9455} & .2869 \\
\multicolumn{1}{c|}{InternLM2-20B-Math} & .4193 & .1165 & .8824 & .2058 \\ \midrule
\multicolumn{5}{c}{\textbf{Instruction   Prompt\_v2}} \\ \midrule
\multicolumn{1}{c|}{GPT-3.5-Turbo} & \underline{.8626} & .2679 & .3529 & .3046 \\
\multicolumn{1}{c|}{GPT-4} & \textbf{.8887} & \textbf{.4309} & \underline{\textbf{.9529}} & \textbf{.5934} \\
\multicolumn{1}{c|}{LLAMA2-70B-Chat} & .6058 & .1709 & .9412 & .2893 \\
\multicolumn{1}{c|}{Mixtral-8*7B-Instruct} & .7886 & \underline{.2727} & .8846 & \underline{.4169} \\
\multicolumn{1}{c|}{Qwen1.5-72B-Chat} & .7643 & .2596 & \underline{\textbf{.9529}} & .4080 \\
\multicolumn{1}{c|}{InternLM2-20B-Chat} & .6329 & .1800 & .9294 & .3016 \\
\multicolumn{1}{c|}{InternLM2-20B-Math} & .4584 & .1202 & .8471 & .2105 \\ \midrule
\multicolumn{5}{c}{\textbf{Instruction   Prompt\_v2 + Knowledge Interpretation}} \\ \midrule
\multicolumn{1}{c|}{GPT-3.5-Turbo} & \underline{.8325} & .2230 & .3882 & .2833 \\
\multicolumn{1}{c|}{GPT-4} & \textbf{.9097} & \textbf{.4839} & .8824 & \textbf{.6250} \\
\multicolumn{1}{c|}{LLAMA2-70B-Chat} & .5968 & .1691 & \underline{.9529} & .2872 \\
\multicolumn{1}{c|}{Mixtral-8*7B-Instruct} & .8313 & \underline{.3208} & .8718 & \underline{.4690} \\
\multicolumn{1}{c|}{Qwen1.5-72B-Chat} & .7914 & .2857 & \textbf{.9647} & .4408 \\
\multicolumn{1}{c|}{InternLM2-20B-Chat} & .6941 & .2059 & .9059 & .3355 \\
\multicolumn{1}{c|}{InternLM2-20B-Math} & .4985 & .1179 & .7529 & .2039 \\ \bottomrule
\end{tabular}}
\end{table}

\subsection{Demonstration Selecting Results}

\label{sec:fewshot}

\begin{table}[!btph]
\caption{Evaluation result of LLMs under few-shot settings. The number (2) and (4) denote the number of demonstrations used in prompt. The best result is marked with \textbf{bold}, and the second best is marked with \underline{underline}.}
\label{tab:few-shot}
\resizebox{.48\textwidth}{!}{
\begin{tabular}{@{}ccccc@{}}
\toprule
\multicolumn{1}{c|}{\textbf{Model}} & \textbf{Accuracy} & \textbf{Precision} &\ \ \  \textbf{Recall}\ \ \ &\ \ \ \ \ \textbf{F1} \ \ \ \ \\ \midrule
\multicolumn{5}{c}{\textbf{Random   Sample (2)}} \\ \midrule
GPT-3.5-turbo & .6572 & .1916 & .9359 & .3181 \\
GPT-4 & .8401 & \underline{.3426} & \textbf{.9487} & \underline{.5034} \\
LLAMA2-70B-Chat & .5432 & .1205 & .8623 & .2115 \\
Mixtral-8*7B-Instruct & \textbf{.8894} & \textbf{.4094} & .6667 & \textbf{.5073} \\
Qwen1.5-72B-Chat & .8231 & .3207 & \underline{.9374} & .4779 \\
InternLM2-20B-Chat & .6921 & .2063 & .9176 & .3369 \\
InternLM2-20B-Math & \underline{.8546} & .2945 & .5059 & .3723 \\ \midrule
\multicolumn{5}{c}{\textbf{Random   Sample (4)}} \\ \midrule
GPT-3.5-turbo & .6597 & .2166 & \textbf{.9463} & .3525 \\
GPT-4 & \underline{.8543} & \textbf{.3659} & \underline{.9615} & \textbf{.5301} \\
LLAMA2-70B-Chat & .5521 & .1369 & .8790 & .2369 \\
Mixtral-8*7B-Instruct & .8280 & .3163 & .8718 & .4642 \\
Qwen1.5-72B-Chat & .8299 & \underline{.3307} & .9498 & \underline{.4906} \\
InternLM2-20B-Chat & .7013 & .2130 & .9034 & .3447 \\
InternLM2-20B-Math & \textbf{.8671} & .3056 & .5290 & .3874 \\ \midrule
\multicolumn{5}{c}{\textbf{Same   Knowledge Sample (2)}} \\ \midrule
GPT-3.5-turbo & .7820 & .1730 & .4103 & .2434 \\
GPT-4 & \textbf{.9025} & \textbf{.4626} & .8718 & \textbf{.6045} \\
LLAMA2-70B-Chat & .5892 & .1547 & \textbf{.9603} & .2665 \\
Mixtral-8*7B-Instruct & .8083 & \underline{.2954} & .8974 & \underline{.4445} \\
Qwen1.5-72B-Chat & .7927 & .2913 & \underline{.9242} & .4430 \\
InternLM2-20B-Chat & .6680 & .1873 & .8563 & .3074 \\
InternLM2-20B-Math & \underline{.8180} & .2532 & .6772 & .3686 \\ \midrule
\multicolumn{5}{c}{\textbf{Match   Sample (2)}} \\ \midrule
GPT-3.5-turbo & \underline{.8883} & .3537 & .3718 & .3625 \\
GPT-4 & .8850 & \underline{.4182} & .8846 & \textbf{.5679} \\
LLAMA2-70B-Chat & .6375 & .2083 & \textbf{.9420} & .3412 \\
Mixtral-8*7B-Instruct & \textbf{.8916} & \textbf{.4286} & .8077 & \underline{.5600} \\
Qwen1.5-72B-Chat & .8587 & .3401 & .8913 & .4923 \\
InternLM2-20B-Chat & .7244 & .2310 & \underline{.9055} & .3681 \\
InternLM2-20B-Math & .8792 & .3476 & .6413 & .4508 \\ \midrule
\multicolumn{5}{c}{\textbf{Mismatch   Sample (2)}} \\ \midrule
GPT-3.5-turbo & .6166 & .1731 & \underline{.9231} & .2915 \\
GPT-4 & \textbf{.8346} & \textbf{.3333} & \textbf{.9359} & \textbf{.4915} \\
LLAMA2-70B-Chat & .4774 & .1029 & .7837 & .1819 \\
Mixtral-8*7B-Instruct & .7831 & \underline{.2674} & .8846 & \underline{.4107} \\
Qwen1.5-72B-Chat & .7363 & .2043 & .8473 & .3292 \\
InternLM2-20B-Chat & .6341 & .1476 & .8207 & .2502 \\
InternLM2-20B-Math & \underline{.7832} & .2471 & .4470 & .3183 \\ \midrule
\multicolumn{5}{c}{\textbf{Diverse   Question Sample (2)}} \\ \midrule
GPT-3.5-turbo & .6613 & .2104 & \underline{.9057} & .3415 \\
GPT-4 & .8576 & \underline{.3700} & \textbf{.9487} & \underline{.5324} \\
LLAMA2-70B-Chat & .5251 & .1104 & .8379 & .1951 \\
Mixtral-8*7B-Instruct & \textbf{.8664} & \textbf{.3778} & .8718 & \textbf{.5272} \\
Qwen1.5-72B-Chat & .8032 & .2951 & .8573 & .4391 \\
InternLM2-20B-Chat & .6785 & .1830 & .8798 & .3030 \\
InternLM2-20B-Math & \underline{.8634} & .3176 & .5145 & .3928 \\ \bottomrule
\end{tabular}}
\end{table}

Based on the conclusion we draw in Sec~\ref{sec:zero-shot}, we choose to use $\mathrm{Instruction Prompt\_v2}$ with knowledge interpretations as the basic instructions for the following few-shot learning experiment. In order to present the different influences brought by each demonstration selection strategy, we use the random sampled demonstrations as the baselines for all the following comparisons. The complete evaluation results of different demonstration selecting strategies are shown in Table~\ref{tab:few-shot}. From the comparisons, we can draw the following conclusions: (1) the random sampled demonstrations have different effects to LLMs. For advanced LLM like GPT-4, which achieves satisfying performance even in the zero-shot scenario, the few-shot learning samples may bring negative influence. On the other hand, for other LLMs, such as Mixtral-8*7-Instruct and InternLM2-20B-Math, those demonstrations bring a significant boost to their overall performances. (2) The number of demonstration samples brings marginal beneficial influence to LLMs' performance; (3) The knowledge relevance strategy does not help, and it even causes great degradation in the performance of several LLMs, e.g., Mixtral-8*7B-Instruct; (4) The label distribution of demonstrations plays an important role in the few-shot learning scenario. From the results, we can observe a consistent improvement in all LLMs by using the all-matching demonstrations. Meanwhile, the all-mismatching samples selecting strategy cause LLMs receive a significant drop in performance. Based on these facts, we draw the conclusion that the matching samples are more informative than the mismatching ones during the few-shot learning scenario, and it will be more efficient to teach LLM with correct samples compared to the wrong cases; (5) the diversity selecting strategy does not behave consistently with different LLMs. Overall, GPT-4 is still the best LLMs, which receives the most first place results. Mixtral-8*7B-Instruct behave surprisingly well with the help of few-shot learning demonstrations. InternLM2-20B-Math is another LLM which receive huge performance boost from the few-shot learning samples and it consistently outperform InternLM2-20B-Chat demonsrate the importance in domain-knowledge for knowledge tagging problems.

\subsection{Self-Reflection Results}

\begin{figure}[!btph]
    \centering
    \includegraphics[width=0.48\textwidth]{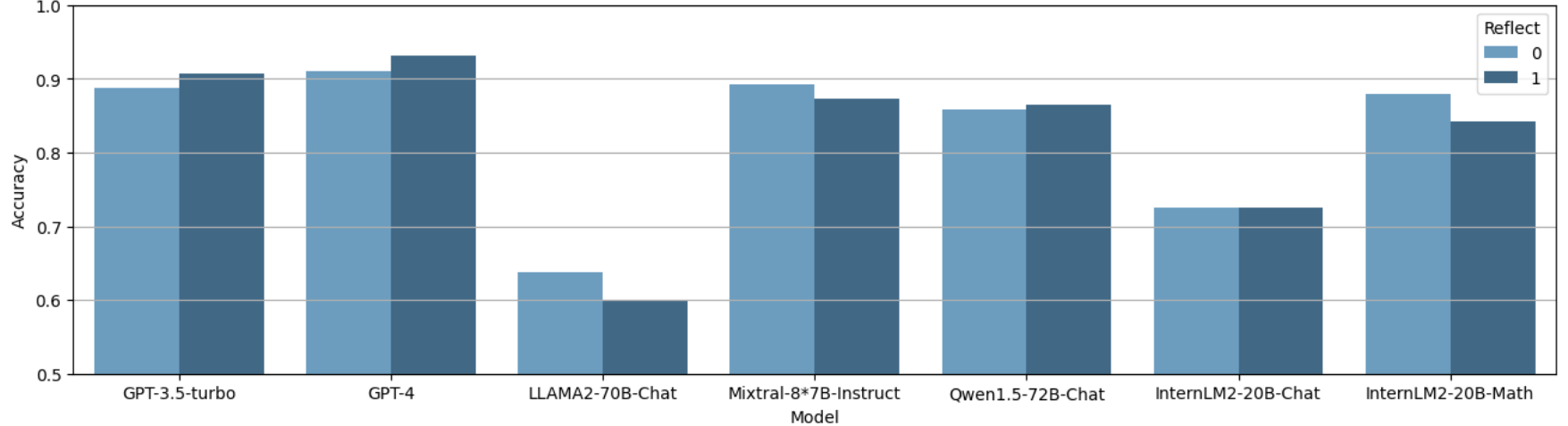}
    \hfill
    \includegraphics[width=0.48\textwidth]{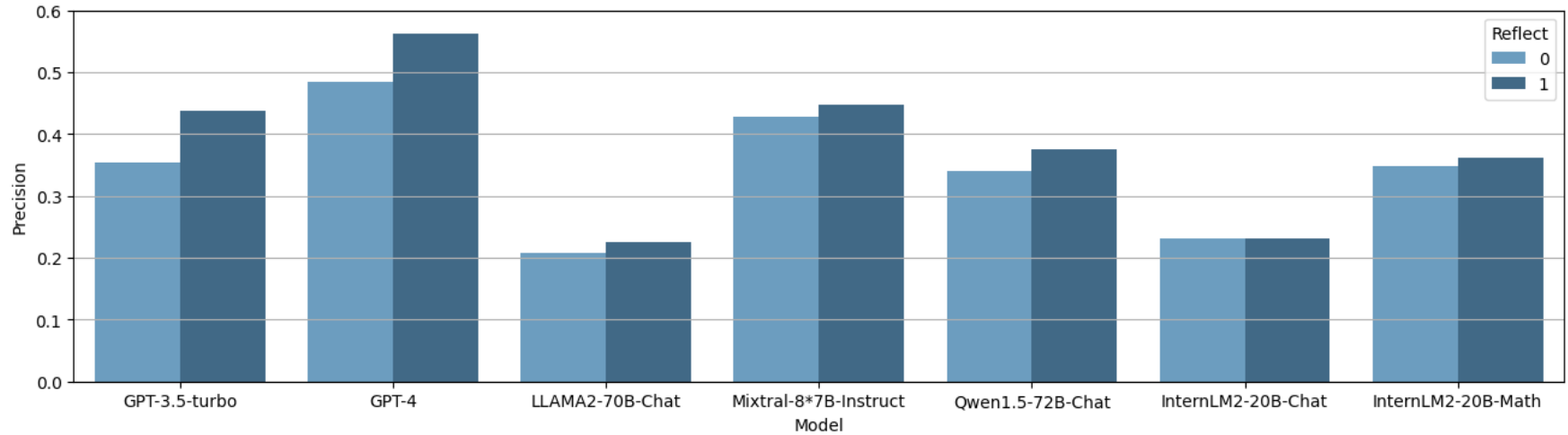}
    \hfill
    \includegraphics[width=0.48\textwidth]{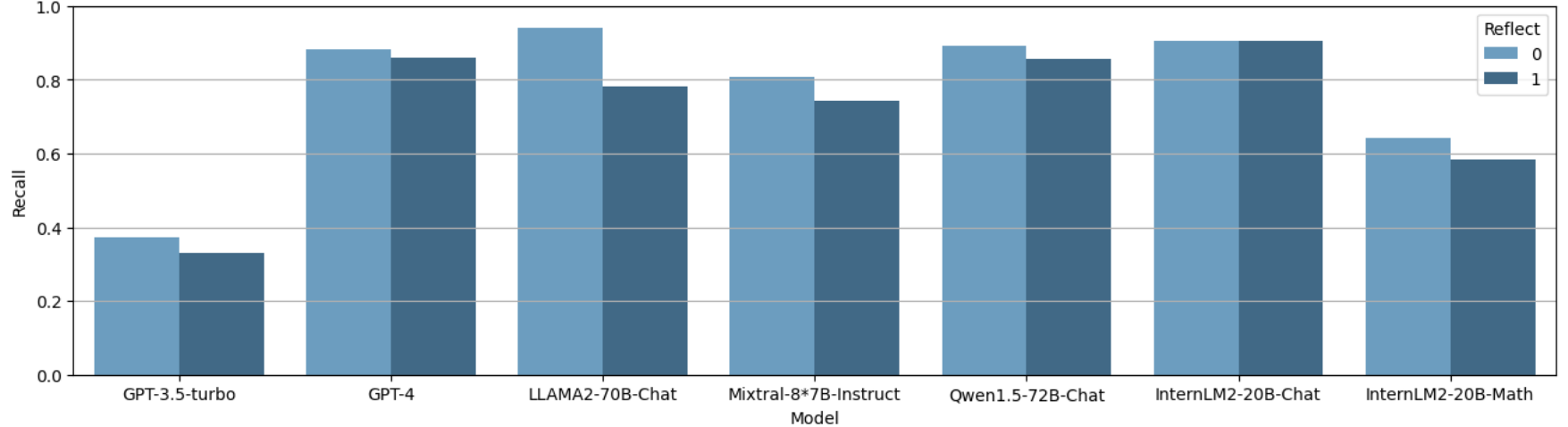}
    \hfill
    \includegraphics[width=0.48\textwidth]{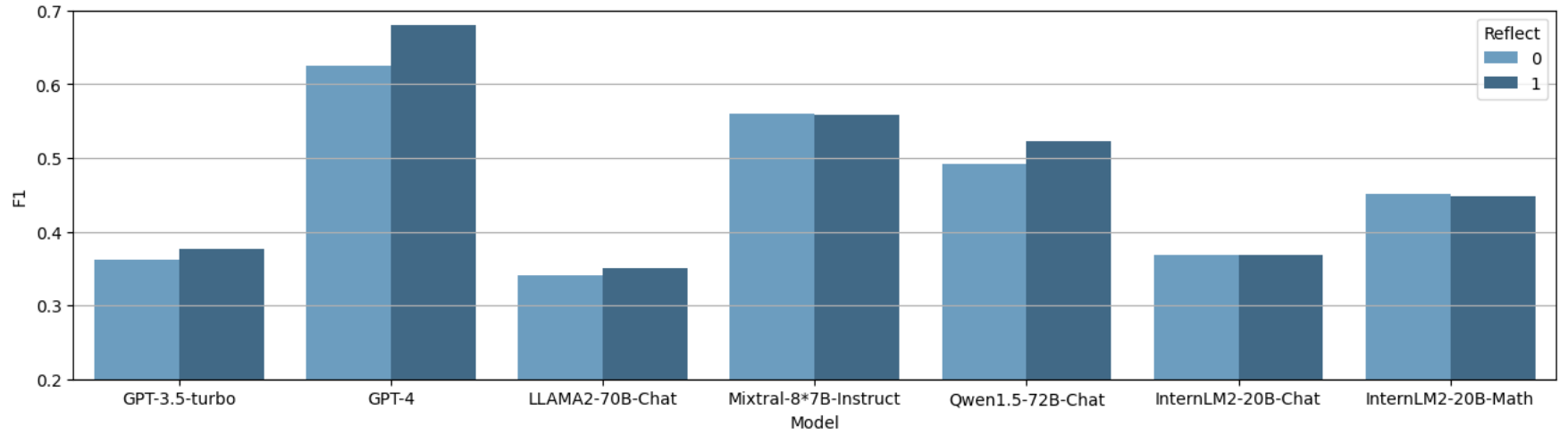}
    \caption{Evaluation result of LLMs with self-reflection revisions, where 0 denotes non-reflect and 1 denotes reflect.}
    \label{fig:self-reflect}
\end{figure}

In the last experiment, we evaluated the effectiveness of self-reflection using the optimal prompt and demonstration settings identified in the preceding sections. The contrasting results between w/ non-reflective and w/o reflective processes are presented in Figure~\ref{fig:self-reflect}. From the figure,  we can observe that incorporating a self-reflection step significantly improves the precision of outcomes generated by various LLMs. This result is aligning with our objective of achieving high precision in knowledge concept tagging tasks. Notably, the extent of improvement varies across different LLMs, where more advanced models, such as GPT-4, show greater benefits from reflection, whereas less advanced ones, like LLAMA2-70B-Chat and InternLM2-20B-Chat, exhibit minimal performance changes. This observation aligns with findings from previous research \cite{zhao2023expel}. Importantly, since self-reflection was applied only to samples with a first match response, there was a potential decrease in recall. However, evaluation of the balanced recall-precision metric, the F1 score, indicates that the self-reflection step enhances performance across all LLMs. Consequently, we conclude that self-reflection is a beneficial process for implementing LLMs in knowledge concept tagging tasks.

\section{Conclusion}

In this paper, we investigate how to leverage large language models to automate knowledge concept tagging for math questions, which is a crucial aspect of intelligent and personalized educational systems. By utilizing the mathematical and logical inference capabilities of LLMs, we show effectiveness in this domain without requiring extensive expert annotations. Our experiments with various LLMs demonstrate promising performance and underscore the importance of prompt design and knowledge interpretation. Our research contributes to the integration of LLMs into education, offering an effective solution for automating knowledge-intensive tasks in education applications.

\appendix
\section{Zero-shot Examples}
We present the judgement results of representative LLMs, including GPT-4, LLAMA2 and Mixtral in Table~\ref{tab:data_sample}. From the table, we cane observe the characteristics of the answers generated by different LLMs. For example, GPT-4's answers are usually precise and accurate and LLAMA2's replies are always verbose and contains some hallucination results. 

\begin{table*}[]
\caption{Sample predictions of LLMs. The wrong results are marked with the \textcolor{red}{Red} color.}
\label{tab:data_sample}
\resizebox{.95\textwidth}{!}{
\begin{tabular}{@{}p{3cm}p{4cm}p{5cm}p{8cm}p{7cm}@{}}
\toprule
Knowledge & Question & GPT-4 & LLAMA-2-70B-Chat & Mixtral-8x7B-Instruct \\ \midrule
Adding and subtracting integer tens - Comparison of formulas & To build a road, Team A will complete it in 8 days alone, and Team B will complete it in 7 days. The   working time ratio of Team A and Team B is \_\_\_\_, and the work efficiency   ratio is \_\_\_\_. & No, the knowledge provided does   not match the question. The question is discussing a problem related to work   efficiency and time management between two teams working on a project, while   the knowledge provided is about mathematical calculations and operations   involving whole tens. & No, the given Knowledge does not   match the Question.The Knowledge provided is about comparing the result of   adding two whole tens with another number, or comparing the size of   expressions involving the addition or subtraction of two whole tens, which   requires a one-step addition or subtraction of two whole tens. However, the   Question is about finding the working time ratio and work efficiency ratio of   two teams, which is not related to adding or subtracting whole tens.   Therefore, the Knowledge does not match the Question. & No, the given knowledge is not   matching the question. The knowledge is about comparing the result of adding   two whole tens with another number or comparing the size of expressions   involving the addition or subtraction of two whole tens. However, the question   is asking about the working time ratio and work efficiency ratio of two teams   working on building a road. The question covers other knowledge that is not   related to the given knowledge. \\ \midrule
Area   unit--unit conversion & If a bag of dolls is divided   equally among 4 children, 3 are left. If it is divided equally among 8   children, 1 is missing. If it is divided equally among 10 children, 1 is   missing. This bag of dolls has at least \_ \_\_\_indivual. & No, the knowledge does not match   the question. The question is about simple arithmetic and finding the least   common multiple, while the knowledge provided pertains to unit conversion   between different area units, which is not relevant to the question. & No, the knowledge does not match   the question. The question is asking for a numerical answer to a word   problem, whereas the knowledge provided is about unit conversion between   different units of area. The knowledge does not include word problems or   provide any information that can be used to solve the problem in the   question. Therefore, it is not possible to determine the number of individual   dolls in the bag using the provided knowledge. & No, the given knowledge is not   matching the question. The question is about dividing a bag of dolls equally   among children and determining the minimum number of dolls in the bag, while   the knowledge is about unit conversion between square meters, square decimeters,   square centimeters, and square millimeters. \\ \midrule
Comparison   of numbers within 5 & Written calculation:   8.49-2.45=\_\_\_\_ & No, the knowledge does not match   the question. While the knowledge talks about comparison between integer   numbers within 5, the question is asking for a calculation involving decimal   numbers. & No, this question does not match   the given knowledge. The question asks for a written calculation, which is   not within the scope of the provided knowledge. The knowledge only covers   comparisons between integer numbers, and does not include calculations. Therefore,   this question cannot be answered using the given knowledge. & No, the knowledge provided is   not matching the question. The question requires a written calculation to   determine the difference between two decimal numbers, while the knowledge   provided is about comparing integer numbers within 5. \\ \midrule
The   composition of numbers within 20 & The adjacent numbers of a number   consisting of 1 tens and 9 ones are \_\_\_\_ and \_\_\_\_. (Case from smallest to   uppercase) & Yes, the knowledge matches the   question. The question asks about identifying the adjacent numbers of a two   digit number, which is formed from 1 tens and 9 ones. This fits within the   knowledge scope of giving several tens and ones and filling in the corresponding   two-digit number. & Yes, the knowledge matches the   question. The question asks for the adjacent numbers of a number consisting   of 1 tens and 9 ones, which is 19. The knowledge states that the number 19   can be broken down into 1 tens and 9 ones. Therefore, the adjacent numbers of   19 are 18 and 20, which answers the question. & \textcolor{red}{No, the knowledge does not match   the question. The question asks for the adjacent numbers of a specific number   (1 ten and 9 ones), while the knowledge is about directly giving a two-digit   number and reading out how many tens or ones it contains, or giving several   tens and several ones and filling in the corresponding two-digit number. The   knowledge does not provide any information about the adjacent numbers of a   specific number.} \\ \midrule
The   rules of decimal point movement-calculation & The bookstore has newly shipped   100 storybooks, with a total thickness of 1.32m. One such storybook is \_\_\_\_m   thick, and 10,000 such storybooks are \_\_\_\_m thick. & No, the knowledge does not match   the question. The question is about finding the thickness of one storybook   and then calculating the total thickness of 10,000 such storybooks, which   involves division and multiplication respectively. However, the knowledge   specifically refers to moving decimal points when multiplying or dividing by   10, 100, or 1000, rather than an application of division and multiplication   in a context. & \textcolor{red}{Yes, the knowledge matches the   question. The question asks for the thickness of one storybook and 10,000   storybooks. The knowledge provides a method to calculate the thickness of a   decimal by multiplying or dividing it by 10, 100, or 1000. To answer the question,   we can use the method provided in the knowledge to calculate the thickness of   one storybook and 10,000 storybooks. For example, to find the thickness of   one storybook, we can multiply 1.32m by 1000 (since there are 1000 storybooks   in the shipment) to get 1320mm. To find the thickness of 10,000 storybooks,   we can divide 1320mm by 1000 to get 1.32m. Therefore, one storybook is 1320mm   thick, and 10,000 storybooks are 1.32m thick. The knowledge provides the   necessary information to answer the question, so the answer is} & No, the knowledge provided does   not match the question. The knowledge is about the movement of the decimal   point when multiplying or dividing decimals by 10, 100, or 1000. However, the   question is asking about the thickness of a storybook and the total thickness   of 10,000 storybooks. The knowledge provided does not give any information   about the thickness of a single storybook or the total thickness of 10,000   storybooks. Therefore, the knowledge cannot be used to answer the question. \\ \midrule
Two-step   operation of adding and subtracting integer tens & Use vertical calculation:   303÷10=\_\_\_\_……\_\_\_\_ & No, the knowledge does not match   the question. The question is asking for a division calculation to be   performed, specifically dividing 303 by 10. However, the knowledge provided   only covers the topics of addition and subtraction of whole tens, not division. & No, the question does not match   the given knowledge. The question asks for a division problem, while the   knowledge provided is about adding and subtracting whole tens. The knowledge   does not cover division, and therefore, it cannot be used to answer the question. & No, the knowledge does not match   the question. The knowledge is about adding and subtracting whole tens, while   the question is about dividing a two-digit number by ten. \\ \midrule
Understanding   and Classification of Numbers & There are \_\_\_\_ ten thousand and   \_\_\_\_ ones in 70200090. & No, the knowledge does not match   the question. The question is asking about place value in a specific number   which is not covered in the knowledge statement. The knowledge provided   pertains to the understanding of various types of numbers, their definitions,   and the classification of these numbers. However, it does not cover concepts   related to the decomposition of numbers into place values. & \textcolor{red}{Yes, the knowledge matches the   question. The question asks about the number of ten thousands and ones in a   given number, which falls under the category of understanding numbers and   their classification. The knowledge specified mentions the definition of integers,   which includes the concept of ten thousands and ones. Additionally, the   knowledge mentions the ability to select numbers of a specified type from a   set of numbers and determine whether a number falls within a defined range,   which are both relevant to answering the question. Therefore, the knowledge   provided matches the question and would be useful in answering it correctly.} & \textcolor{red}{Yes, the Knowledge does match   the Question. The Knowledge includes understanding of numbers, including the   definitions of integers and positive numbers. The Question is asking about   the classification of numbers, specifically the place value of a given number.} \\ \midrule
Use   the properties of decimals to rewrite decimals & There are \_\_\_\_ 0.01s in 0.69;   rewriting 0.3 into two decimal places with the same size is \_\_\_\_. & Yes, the knowledge matches the   question. The question addresses two aspects of the provided knowledge: the   conversion of decimals to units of one-hundredth (there are \_\_\_\_ 0.01s in   0.69) and the rewriting of decimals to have a specific number of decimal places   (rewriting 0.3 into two decimal places). Both of these tasks require   understanding and applying the properties of decimals. & Yes, the knowledge matches the   question. The knowledge states that we can rewrite decimals as required by   adding or removing zeroes from the end, and this will not change their value.   The question asks us to rewrite 0.3 into two decimal places with the same   size, which can be done by adding zeroes to the end, making it 0.30.   Therefore, the answer to the question is Yes, the knowledge matches the   question. & \textcolor{red}{No, the question does not match   the knowledge provided. The knowledge is about adding or removing zeros from   the end of a decimal number to change its representation, while the question   asks about the number of tenths and hundredths in a decimal number and   rewriting another decimal number to a specific number of decimal places.} \\ \midrule
Use   the properties of decimals to simplify decimals & The product of two prime numbers   is 35 and the difference is 2. The smaller of the two numbers is \_\_\_\_ and the   larger number is \_\_\_\_. & No, the knowledge does not match   the question. The knowledge is about simplifying decimals, while the question   is about finding two prime numbers that multiply to 35 and have a difference   of 2. This question does not require knowledge about decimals or their   properties. & No, the given Knowledge does not   match the Question. The Question is asking for the values of two prime   numbers, while the Knowledge is about simplifying decimals. The two concepts   are unrelated, and the Knowledge does not provide any information that can   help answer the Question. Therefore, the answer is No. & No, the given knowledge is not   relevant to the question. The question asks about the properties of prime   numbers and their product, while the knowledge is about simplifying decimals. \\ \midrule
reciprocal & Known: \(\frac{6}{{13}}x =   \frac{4}{5}y = 1\), then \(6x + 4y = \)\_\_\_\_ & No, the knowledge does not match   the question. The knowledge talks about writing the reciprocal of a number   and calculating the value of a number which yields the product as 1. However,   the question involves solving equations with fractions and includes addition   of terms to find the answer. & \textcolor{red}{Yes, the knowledge matches the   question. The question asks for the value of \(6x + 4y\) when \(\frac{6}{13}x   = \frac{4}{5}y = 1\). Using the given information, we can write: \(6x + 4y =   6\left(\frac{13}{6}\right) + 4\left(\frac{5}{4}\right) = 13 + 5 = 18\)   Therefore, the answer is 18.} & No, the knowledge provided does   not directly give the value of \(6x + 4y\). The knowledge states that the   reciprocal of a fraction, integer, or decimal should be directly written, or   the number whose product with a given number is 1 should be found (with no   letters involved). However, the question asks for the value of an expression   based on the given equation. The knowledge does not provide enough   information to directly determine the value of \(6x + 4y\) from the given   equation. \\ \bottomrule
\end{tabular}}
\end{table*}

\section{Reflective Examples}

We present the reflective judgment results of GPT-4 in Table~\ref{tab:data_reflect}. From the table, we find that given the prior generated judgment reasons, GPT-4 can discern the wrong concepts and conclusions generated within the intermediate deduction process and fix them automatically. Contributing to this behavior, the model's precision performance gets boosted by a great margin.

\begin{table*}[]
\caption{Reflection predictions generated by GPT-4. The discovered error in original results are marked with the \textcolor{red}{Red} color.}
\label{tab:data_reflect}
\resizebox{.95\textwidth}{!}{
\begin{tabular}{@{}p{4cm}p{4cm}p{9cm}p{9cm}@{}}
\toprule
Knowledge & Problem & Origin Result & Reflection Result \\ \midrule
Comparison   of numbers within 5 & Fill in “>”, “<” or “=\&quot;   on the horizontal lines. 37800\_\_\_\_37,800; 6590000\_\_\_\_659,000 & Yes, the knowledge matches the   question. The question is asking for a comparison between numerical values   without including any calculations. In accordance with the knowledge, the   mentioned numbers are presented and require assigning of comparative symbols. & No, the prior answer is   incorrect. Although the question does involve comparison of numerical values, it doesn't match the knowledge because the knowledge specifically states the   comparison is between integer numbers smaller than 5. \textcolor{red}{However, the numbers in   the question are far larger than 5.} \\ \midrule
Use   the properties of decimals to rewrite decimals & 79 0.1\&\#39;s are \_\_\_\_, and   600 0.001\&\#39;s are \_\_\_\_. (fill in decimals) & Yes, the question matches the   knowledge. The question focuses on converting specific quantities of decimals   to larger or smaller decimal amounts, adhering to the concept of decimal   precision requirements. There is no explicit arithmetic calculation involved,   so it complies with the conditions given in the knowledge. & No, the prior answer is   incorrect. The question \textcolor{red}{does involve explicit arithmetic calculation} as it   asks for the multiplication of 79 by 0.1 and 600 by 0.001, which isn't   consistent with the condition mentioned in the knowledge. \\ \midrule
Adding   and subtracting integer tens - Comparison of formulas & Fill in \&quot;>\&quot;,   \&quot;<\&quot; or \&quot;=\&quot;. 18+7\_\_\_\_33 67-8\_\_\_\_59 & Yes, the knowledge matches the   question. The question involves comparing two expressions involving addition   and subtraction of two digit numbers. And there is a comparing step as asked   in the question. & The prior answer is wrong.   Although the question involves comparing two expressions, it does not follow   the knowledge provided because \textcolor{red}{it includes numbers that are not multiples of tens}. For example, '18+7' and '67-8' are not composed of two-digit multiples   of tens. Therefore, the knowledge does not match with the question. \\ \midrule
Use   the properties of decimals to rewrite decimals & Rewrite 128091 as a number using   \&quot;ten thousand\&quot; as the unit, which is \_\_\_\_ ten thousand, and   then keep three decimal places to get \_\_\_\_ ten thousand. & Yes, the knowledge matches the   question. The question refers to converting a given number into a new number   with specific decimal precision, in this case three decimal places, and the   unit used for conversion is "ten thousand". The knowledge explains   exactly this process without involving any mention of an explicit arithmetic   calculation. & No, the prior response is   incorrect. The knowledge point describes the process of converting numbers   into new numbers with specific decimal precision requirements (tenth,   hundredth, or thousandth place). However, the question is seeking a conversion of the number into "ten thousands" and then asking to   keep three decimal places. \textcolor{red}{The question involves a place value unit   conversion not covered in the given knowledge}. Therefore, these two do not   match perfectly. \\ \midrule
The   rules of decimal point movement-calculation & Calculation: 14*0.016=\_\_\_\_. & Yes, the knowledge matches the   question. The question is asking for a simple one-step arithmetic calculation   involving a multiplication operation. There are no multiple points movement   involved here and it also involves a multiplication with a decimal number   which is allowed according to the given knowledge. & No, the prior answer is   incorrect. The knowledge does not match the question. Based on the provided   knowledge, it specifically refers to "Calculate arithmetic expression   based on the moving of the decimal point". The question   "14*0.016=\_\_\_\_" is a typical arithmetic operation, \textcolor{red}{but it doesn't involve any 'moving of the decimal point' problem that is mentioned in the knowledge.} \\ \midrule
Adding   and subtracting whole tens - word problems & $58-20$=\_\_\_\_ & Yes, the question matches the   knowledge. This question is a one-step application problem involving the   subtraction of whole tens from whole tens. Both numbers, 58 and 20, are   two-digit multiples of tens. And the calculated result, 38, is less than 100.   There is no comparing step in this question, so it does not cover other   knowledge. & No, the prior answer is   incorrect. The numbers in the question, \textcolor{red}{58 and 20, are not multiples of tens}. The problem is a simple subtraction operation rather than a scenario   involving the subtraction of whole tens from whole tens. \\ \midrule
Adding   and subtracting whole tens - word problems & There are 52 ducks and 38 geese   in the pond. How many more ducks than geese are there in the pond? Column   expression and calculation: \_\_\_\_ (only) & Yes, the knowledge matches the   question. The question involves the subtraction of whole tens from whole   tens, and the calculation result is less than 100. The question also has a   comparing step, which is comparing the number of ducks to the number of geese.   Moreover, although the numbers involved are not multiples of tens, they are   two-digit numbers, which is consistent with the specification in the   knowledge that there are only two-digit numbers in expressions. & No, the prior response is not   correct. The knowledge pertains to the addition or subtraction of multiples   of tens only, \textcolor{red}{while the problem in the question uses the numbers 52 and 38,   which are not multiples of tens}. Therefore, the knowledge does not match the   question. \\ \midrule
Area   unit--unit conversion & 420,000 square centimeters =   \_\_\_\_ square meters 64 meters = \_\_\_\_ centimeters & Yes, the knowledge matches the   question because the question is asking for the conversion of area units   between square centimeters and square meters, and length units between meters   and centimeters, which falls under the provided knowledge concerning conversion   between these units. & No, the prior response is   incorrect. \textcolor{red}{The knowledge given only pertains to the conversions of area units, and does not cover the conversion of length units}. Therefore, the portion of the question asking to convert 64 meters to centimeters does not match the given knowledge. Only the part of the question asking to convert   420,000 square centimeters to square meters matches the knowledge. \\ \midrule
Understanding   and Classification of Numbers & The difference between a natural   number and its reciprocal is \&lt;50\textbackslash{}frac\{\{50\}\}\{\{51\}\}\&gt;. This   natural number is \_\_\_\_. & Yes, the knowledge matches the   question. The question is asking to identify a natural number based on a   given condition. The knowledge point refers to understanding definitions of   numbers including natural numbers and involves judging propositions related   to the classification of numbers. This pertains to the content of the   question. & No, the prior answer is   incorrect. While the question does involve the understanding of numbers and   natural numbers, which is part of the knowledge, \textcolor{red}{it also involves number operations to solve the difference between a natural number and its reciprocal which is explicitly stated as not included in the provided   knowledge}. \\ \midrule
Adding   and subtracting integer tens - Comparison of formulas & In the following calculations,   the numbers less than 40 are \_\_\_\_ (fill in the letters in order, connected   with commas) A: 25+14; B: 68-27; C: 13+36; D: 83-52 & Yes, the given knowledge matches   the question. The question has a comparing step which is to find the numbers   less than 40. It involves only addition or subtraction between two-digit   numbers which can be multiples of tens. & No, the prior answer is incorrect. \textcolor{red}{The question involves operations with numbers that are not multiples of tens}, contrary to the requirement of the Knowledge that only   multiples of tens should be involved. \\ \midrule
Adding   and subtracting integer tens - Comparison of formulas & In the following formula, there   are \_\_\_\_ numbers that are more than eighty: A: 8 + 21; B: 46 + 35; C: 92-5;   D: 95-14 & Yes, the knowledge is matching   with the question. The question is about comparing expression results   (involving addition or subtraction) with an external number 80. The   expressions mentioned involves only two digits and all numbers involved are   multiples of tens. Also, the question process involves a comparison step,   which fits with the knowledge point. & The prior answer is wrong. The   knowledge does not match the question. The knowledge specifies that the expressions should involve only two-digit multiples of tens. In the question, \textcolor{red}{the numbers used are not all multiples of tens}. For instance, 21, 5, and 14 are used, and these are not multiples of tens. Hence, the question does not   align fully with the knowledge. \\ \bottomrule
\end{tabular}}
\end{table*}

\bibliographystyle{named}
\bibliography{ijcai24}

\end{document}